\pdfoutput=1

\documentclass[11pt,a4paper]{article}
\usepackage[hyperref]{emnlp2020}
\usepackage{times}
\usepackage{latexsym}

\usepackage{microtype}

\aclfinalcopy %
\def\aclpaperid{***} %

\usepackage{graphicx} %
\PassOptionsToPackage{hyphens}{url}\usepackage{url}
\usepackage{enumitem} %
\usepackage{multirow} %
\usepackage{amsthm} %
\usepackage{algorithm,algorithmic} %
\usepackage{caption} %
\usepackage{xspace} %
\usepackage{soul}

\usepackage{amsmath,amssymb,mathtools,bbm,tikz}
\newcommand{\R}{\mathbb{R}}

\DeclareMathOperator*{\minimize}{minimize}

\def\1{\mathbbm{1}}
\DeclarePairedDelimiter\abs{\lvert}{\rvert}%
\newcommand{\para}[1]{\left({#1}\right)}

\newcommand{\curlypara}[1]{\left\{{#1}\right\}}
\def\cpara{\curlypara}

\renewcommand{\vec}[1]{\boldsymbol{#1}}

\newcommand{\subword}{\subseteq}

\newcommand{\code}[1]{{\ttfamily{#1}}}
\usepackage{seqsplit}

\newcommand{\bos}{{BoS}\@\xspace}
\newcommand{\pbos}{{PBoS}\@\xspace}
\newcommand{\pbosn}{{PBoS-n}\@\xspace}

\newtheorem{proposition}{Proposition}

\newenvironment{example}{\begin{proof}}{\end{proof}}

\newcommand{\algref}[1]{Algorithm~\ref{#1}}
\newcommand{\eqnref}[1]{Eq.~\eqref{#1}}
\newcommand{\figref}[1]{Figure~\ref{#1}}
\newcommand{\secref}[1]{Section~\ref{#1}}
\newcommand{\tabref}[1]{Table~\ref{#1}}

\newif\ifshowtodos
\showtodosfalse
\ifshowtodos
\newcommand{\comment}[3]{{\color{#1}{\noindent\textit{\textbf{#2}: {#3}}}}}
\newcommand{\yingyu}[1]{\comment{red}{Yingyu}{#1}}
\newcommand{\shawn}[1]{\comment{red}{Shawn}{#1}}
\newcommand{\jinman}[1]{\comment{purple}{Jinman}{#1}}
\newcommand{\xiaomin}[1]{\comment{cyan}{xiaomin}{#1}}
\newcommand{\todo}[1]{\comment{red}{todo}{#1}}
\else
\newcommand{\yingyu}[1]{}
\newcommand{\shawn}[1]{}
\newcommand{\jinman}[1]{}
\newcommand{\xiaomin}[1]{}
\newcommand{\todo}[1]{}
\fi

\title{{PBoS}: Probabilistic Bag-of-Subwords for Generalizing Word Embedding}

\author{%
  Zhao Jinman \; Shawn Zhong \; Xiaomin Zhang \; Yingyu Liang \\
  University of Wisconsin-Madison, Madison, WI, USA \\
  \texttt{\{jz,yliang\}@cs.wisc.edu} \\
  \texttt{\{shawn.zhong,xzhang682\}@wisc.edu} \\
}

\date{}

\begin{document}
\maketitle
\begin{abstract}
  We look into the task of \emph{generalizing} word embeddings: given a set of pre-trained word vectors over a finite vocabulary, the goal is to predict embedding vectors for out-of-vocabulary words, \emph{without} extra contextual information.
  We rely solely on the spellings of words and propose a model, along with an efficient algorithm, that simultaneously models subword segmentation and computes subword-based compositional word embedding.
  We call the model probabilistic bag-of-subwords (PBoS), as it applies bag-of-subwords for all possible segmentations based on their likelihood.
  Inspections and affix prediction experiment show that PBoS is able to produce meaningful subword segmentations and subword rankings without any source of explicit morphological knowledge.
  Word similarity and POS tagging experiments show clear advantages of PBoS over previous subword-level models in the quality of generated word embeddings across languages.
\end{abstract}

\section{Introduction}
\label{sec:introduction}
Word embeddings pre-trained over large texts have demonstrated benefits for many NLP tasks, especially when the task is label-deprived.
However, many popular pre-trained sets of word embeddings assume fixed finite-size vocabularies~\footnote{\url{https://code.google.com/archive/p/word2vec/},  \citet{mikolov2013distributed-w2v}.}\textsuperscript{,}~\footnote{\url{https://nlp.stanford.edu/projects/glove/}, \citet{pennington2014glove}.}, which hinders their ability to provide useful word representations for out-of-vocabulary (OOV) words.

We look into the task of \emph{generalizing} word embeddings: extrapolating a set of pre-trained word embeddings to words out of its fixed vocabulary, \emph{without} extra access to contextual information (e.g. example sentences or text corpus).
In contrast, the more common task of \emph{learning} word embeddings, or often just \emph{word embedding}, is to obtain distributed representations of words directly from large unlabeled text.
The motivation here is to extend the usefulness of pre-trained embeddings without expensive retraining over large text.

There have been works showing that contextual information can also help generalize word embeddings \citep[for example, ][]{khodak2018carte, schick2019attentive, schick2019learning}. 
We here, however, focus more on the research question of how much one can achieve from just word compositions. In addition, our proposed way of utilizing word composition information can be combined with the contextual embedding algorithms to further improve the performance of generalized embeddings.

The hidden assumption here is that words are made of meaningful parts (cf. morphemes) and that the meaning of a word is related to the meaning of their parts.
This way, humans are often able to guess the meaning of a word or term they have never seen before.
For example, ``postEMNLP'' probably means ``after EMNLP''.

Different models have been proposed for that task of generalizing word embeddings using word compositions, usually under the name of subword(-level) models. \citet{stratos2017reconstruction-rnn, pinter2017mimicking, kim2018learning-cnn} model words at the character level. However, they have been surpassed by later subword-level models, probably because of putting too much burden on the models to form and discover meaningful subwords from characters.
Bag-of-subwords (\bos) is a simple yet effective model for learning~\citep{bojanowski2017enriching-fasttext} and generalizing~\citep{zhao2018generalizing-bos} word embeddings. 
\bos composes a word embedding vector by taking the sum or average of the vectors of the subwords (character $n$-grams) that appear in the given word. 
However, it ignores the importance of different subwords since all of them are given the same weight.
Intuitively, ``farm'' and ``land'' should be more relevant in composing representation for word ``farmland'' than some random subwords like ``armla''.

Even more favorable would be a model's ability to discover meaningful subword segmentations on its own.
\citet{cotterell2016morphological} bases their model over morphemes but needs help from an external morphological analyzer such as Morfessor~\citep{virpioja2013morfessor}. 
\citet{sasaki2019subword} use trainable self-attention to combine subword vectors. 
While the attention implicitly facilitates interactions among subwords, there has been no explicit enforcement of mutual exclusiveness from subword segmentation, making it sometimes difficult to rule out less relevant subwords.
For example, ``her'' is itself a likely subword, but is unlikely to be relevant for ``higher'' as the remaining ``hig'' is unlikely.

We propose the probabilistic bag-of-subwords (\pbos) model for generalizing word embedding.
\pbos simultaneously models subword segmentation and composition of word representations out of subword representations.
The subword segmentation part is a probabilistic model capable of handling ambiguity of subword boundaries and ranking possible segmentations based on their overall likelihood.
For each segmentation, we compose a word vector as the sum of all subwords that appear in the segmentation.
The final embedding vector is the expectation of the word vectors from all possible segmentations.
An alternative view is that the model assigns word-specific weights to subwords based on how likely they appear as meaningful segments for the given word.
Coupled with an efficient algorithm, our model is able to compose better word embedding vectors with little computational overhead compared to \bos.

Manual inspections show that \pbos is able to produce subword segmentations and subword weights that align with human intuition.
Affix prediction experiment quantitatively shows that the subword weights given by \pbos are able to recover most eminent affixes of words with good accuracy.

To assess the quality of generated word embeddings, we evaluate with the intrinsic task of word similarity which relates to the semantics; 
as well as the extrinsic task of part-of-speech (POS) tagging which requires rich information to determine each word's role in a sentence.
English word similarity experiment shows that \pbos improves the correlation scores over previous best models under various settings and is the only model that consistently improves over the target pre-trained embeddings.
POS tagging experiment over 23 languages shows that \pbos improves accuracy compared in all but one language to the previous best models, often by a big margin.

\bigskip
We summarize our contributions as follows:
\begin{itemize}[leftmargin=*]
  \setlength\itemsep{-0.5em}
  \item We propose \pbos, a subword-level word embedding model that is based on probabilistic segmentation of words into subwords, the first of its kind (\secref{sec:model}).
  \item We propose an efficient algorithm that leads to an efficient implementation~\footnotemark of \pbos with little overhead over previous much simpler \bos. (\secref{sec:algorithm}).
  \item Manual inspection and affix prediction experiment show that \pbos is able to give reasonable subword segmentations and subword weights (\secref{sec:seg-exp} and \ref{sec:seg-affix-pred}).
  \item Word similarity and POS tagging experiments show that word vectors generated by \pbos have better quality compared to previously proposed models across languages (\secref{sec:word-sim} and \ref{sec:pos-tagging}).
\end{itemize}
\footnotetext{
Code used for this work can be found at \url{https://github.com/jmzhao/pbos}.
}

\section{PBoS Model}
\label{sec:model}

Following the above intuition, in this section we describe the \pbos model in detail.

We first develop a model that segments a word into subword and associates each subword segmentation with a likelihood based on the meaningfulness of each subword segment.
We then apply \bos over each segmentation to compose a ``segmentation vector''. 
The final word embedding vector is then the probabilistic expectation of all the segmentation vectors.
The subword segmentation and likelihood association part require no explicit source of morphological knowledge and are tightly integrated with the word vector composition part, which in turn gives rise to an efficient algorithm that considers \emph{all} possible segmentations simultaneously (\secref{sec:algorithm}).
The model can be trained by fitting a set of pre-trained word embeddings.

\subsection{Terminology}
For a given language, let $\Gamma$ be its alphabet.
A \emph{word} $w$ of length $l = \abs{w}$ is a string made of $l$ letters in $\Gamma$, i.e. $w = c_1 c_2 \dots c_l \in \Gamma^l$ where $w[i] = c_i$ is the $i$-th letter. 
Let $p_w \in [0, 1]$ be the probability that $w$ appears in the language.
Empirically, this is proportional to the unigram frequency of word $w$ observed in large text in that language.

Note that we do not assume a vocabulary. 
That is, we do not distinguish words from arbitrary strings made out of the alphabet.
The implicit assumption here is that a ``word'' in common sense is just a string associated with high probability. 
In this sense, $p_w$ can also be seen as the likelihood of string $w$ being a ``legit word''.
This blurs the boundary between words and non-words, and automatically enables us to handle unseen words, alternative spellings, typos, and nonce words as normal cases.

We say a string $s \in \Gamma^+$ is a \emph{subword} of word $w$, denoted as $s \subword w$,
if $s = w[i:j] = c_i \dots c_j$ for some $1 \le i \le j \le \abs{w}$, i.e. $s$ is a substring of $w$.
The probability that subword $s$ appears in the language can then be defined as 
\begin{equation}
  p_s \propto \sum_{w \in \Gamma^+} p_w \sum_{1 \le i \le j \le \abs{w}} \1 (s = w[i:j])
\end{equation}
where $\1(pred)$ gives 1 and otherwise 0 only if $pred$ holds.
Note that a subword $s$ may occur more than once in the same word $w$.
For example, subword ``ana'' occurs twice in the word ``banana''.

A \emph{subword se\underline{g}mentation} $g$ of word $w$ of length $k = \abs{g}$ is a tuple $(s_1, s_2, \dots, s_k)$ of subwords of $w$, so that $w$ is the concatenation of $s_1, \dots, s_k$.

\subsection{Probabilistic Subword Segmentation}

A subword transition graph for word $w$ is a directed acyclic graph $G_w = \para{N_w, E_w}$.
Let $l=\abs{w}$.
The vertices $N_w = \cpara{0,\dots,l}$ correspond to the positions between $w[i]$ and $w[i+1]$ for all $i \in [l-1]$, as well as to the beginning (vertiex 0) and the end (vertex $l$) of $w$.
Each edge $(i,j) \in E_w = \cpara{(i,j) : 0 \le i < j \le l}$ corresponds to subword $w[i:j]$. 
We use $G_w$ as a useful image for developing our model.

\begin{proposition}
  Paths from $0$ to $\abs{w}$ in $G_w$ are in one-to-one correspondence to segmentations of $w$.
\end{proposition}
\begin{proposition}
  \label{prop:num-seg}
  There are $2^{\abs{w}-1}$ different possible segmentations for word $w$.
\end{proposition}

Each edge $(i, j)$ is associated with a weight $p_{w[i:j]}$ --- how likely $w[i:j]$ itself is a meaningful subword.
We model the likelihood of segmentation $g$ being a segmentation of $w$ as being proportional to the product of all its subword likelihood -- the transition along a path from $0$ to $\abs{w}$ in $G_w$:
\begin{equation}
\label{eqn:prob-seg}
  p_{g \mid w} \propto \prod_{s \in g} p_s.
\end{equation}

\begin{figure}
\centering
\resizebox{\columnwidth}{!}{%
  \begin{tikzpicture}[shorten >=1pt,node distance=2cm,auto]
    \tikzstyle{vertex}=[shape=circle,thick,draw,minimum size=1cm]
    \tikzstyle{vertex_special}=[shape=circle,double,thick,draw,minimum size=1cm]

    \node[vertex_special](0){0};
    \node[vertex,right of=0](1){1};
    \node[vertex,right of=1](2){2};
    \node[vertex,right of=2](3){3};
    \node[vertex,right of=3](4){4};
    \node[vertex,right of=4](5){5};
    \node[vertex_special,right of=5](6){6};

    \newcommand{\edge}[4]{
        (#1)[#4]
        edge
            node [above] {\Large #3}
            node [below] {\Large $p_\text{``#3''}$}
        (#2)
    }

    \path[->]
        \edge{0}{1}{h}{}
        \edge{1}{2}{i}{}
        \edge{2}{3}{g}{}
        \edge{3}{4}{h}{}
        \edge{4}{5}{e}{}
        \edge{5}{6}{r}{}

        \edge{0}{2}{hi}{bend left=60}
        \edge{2}{6}{gher}{bend left=80}
        \edge{2}{4}{gh}{bend left=60}
        \edge{3}{6}{her}{bend right=60}
   ;

   \path[->, very thick]
        \edge{0}{4}{high}{bend right=45}
        \edge{4}{6}{er}{bend left=60}
   ;

  \end{tikzpicture}
}
\caption{\small
  Diagram of probabilistic subwords transitions for word ``higher''.
  Some edges are omitted to reduce clutter.
  Each edge is labeled by a subword $s$ of the word, associated with $p_s$.
  \textbf{Bold} edges constituent a path from node 0 to 6, corresponding to the segmentation of the word into ``high'' and ``er''.
}
\label{fig:prob-transit}
\end{figure}
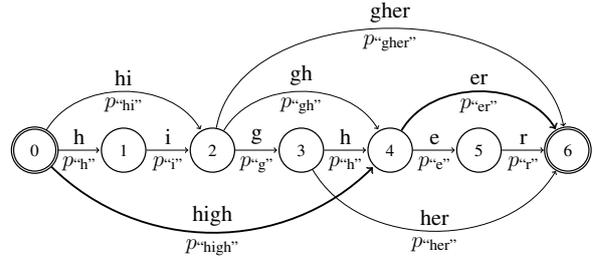

\begin{example}
  \figref{fig:prob-transit} illustrates $G_w$ for word $w = \text{``higher''}$ of length $6$.
  Bold edges $(0, 4)$ and $(4, 6)$ form a path from $0$ to $6$, which corresponds to the segmentation $(\text{``high''}, \text{``er''})$.
  The likelihood $p_{(\text{``high''}, \text{``er''}) \mid w}$ of this particular segmentation is proportional to $p_{\text{``high''}} p_{\text{``er''}}$ -- the product of weights along the path.
\end{example}

\subsection{Probabilistic Bag-of-Subwords}
\label{sec:pseg}
Based on the above modeling of subword segmentations, we propose the Probabilistic Bag-of-Subword (PBoS) model for composing word embeddings. 

The embedding vector $\vec{w}$ for word $w$ is the expectation of all its segmentation-based word embedding:
\begin{equation}
  \label{eqn:sum-seg}
  \vec{w} = \sum_{g \in Seg_w} p_{g|w} \vec{g}
\end{equation}
where $\vec{g}$ is the embedding for segmentation $g$.

Given a subword segmentation $g$, we adopt the Bag-of-Subwords (BoS) model~\cite{bojanowski2017enriching-fasttext,zhao2018generalizing-bos} for composing word embedding from subwords.
Specifically, we apply \bos~\footnote{\citet{zhao2018generalizing-bos} used averaging instead of summation. However, both give uniform weights to all subwords and result in vectors only differ by a scalar factor. We thus do not distinguish the two and refer to either of them as \bos.} over the subword segments in $g$:
\begin{equation}
  \label{eqn:emb-seg}
  \vec{g} = \sum_{s \in g} \vec{s},
\end{equation}
where $\vec{s}$ is the vector representation for subword $s$,
as if the current segmentation $g$ is the ``golden'' segmentation of the word.
In such case, we assume the meaning of the word is the combination of the meaning of all its subword segments.
We maintain a look-up table $S : \Gamma^+ \to \R^d$ for all subword vectors (i.e. $\vec{s} = S(s)$) as trainable parameters of the model, where $d$ is the embedding dimension.

Combining \eqnref{eqn:sum-seg} and \eqref{eqn:emb-seg}, we can compose vector representation for any word $w \in \Gamma^+$ as
\begin{equation}
  \label{eqn:sum-seg-all}
  \vec{w} = \sum_{g \in Seg_w} p_{g \mid w} \sum_{s \in g} \vec{s}  .
\end{equation}

Given a set of target pre-trained word vectors $\vec{w}^*$ defined for words within a finite vocabulary $W$,
our model can be trained by minimizing the mean square loss:
\begin{equation}
    \minimize_{S} \frac{1}{\abs{W}} \sum_{w \in W} \|\vec{w} - \vec{w}^*\|_2^2.
\end{equation}

\section{Efficient Algorithm}
\label{sec:algorithm}
\pbos simultaneously considers all possible subword segmentations and their contributions in composing word representations.
However, summing over embeddings of all possible segmentations can be awfully inefficient,
as simply enumerating all possible segmentations of $w$ takes number of steps exponential to the length of $w$ (Proposition~\ref{prop:num-seg}).
We therefore need an efficient way to compute \eqnref{eqn:sum-seg-all}.

\subsection{Alternative View: Weighted Subwords}
\label{sec:alternative-view}
Exchanging the order of summations in \eqnref{eqn:sum-seg-all} from segmentation first to subword first, we get 
\begin{align}
  \vec{w}
  \label{eqn:sum-subword}
  =& \sum_{s \subword w} a_{s \mid w} \vec{s} 
\end{align}
where
\begin{equation}
  \label{eqn:subword-weight}
  a_{s \mid w} \propto \sum_{g \in Seg_w,\ g \ni s} p_{g \mid w}
\end{equation}
is the weight accumulated over subword $s$, summing over all segmentations of $w$ that contain $s$.~\footnote{For simplicity, here we assume all subwords are unique in $w$. A more careful index-based summation would model the general case but the idea remains the same. We take care of this in Algorithm \ref{alg:subword-weights}.
}

\eqnref{eqn:sum-subword} provides an alternative view of the word vector composed by our model: a weighted sum of all the word's subword vectors.
Comparing to \bos, we assign different importance $a_{s \mid w}$, instead of a uniform weight, to each subword.
$a_{s \mid w}$ can be viewed as the likelihood of subword $s$ being a meaningful segment of the particular word $w$, considering both the likelihood of $s$ itself being meaningful, and at the same time how likely the rest of the word can still be segmented into meaningful subwords.

\begin{example}
  Consider the contribution of subword $s = \text{``gher''}$ in word $w = \text{``higher''}$.
  Possible contributions only come from segmentations that contain ``higher'': $g_1=\ $ (``h'', ``i'', ``gher'') and $g_2=\ $ (``hi'', ``gher'').
  Each segmentation $g$ adds weight $p_{g \mid w}$ to $a_{s \mid w}$.
  In this case, $a_{\text{``gher''} \mid w}$ will be smaller than $a_{\text{``er''} \mid w}$ because both $p_{g_1 \mid w}$ and $p_{g_2 \mid w}$ would be rather small. 
\end{example}

\subsection{Computing Subword Weights}
\label{sec:comp-subword-weights}
Now we can efficiently compute \eqnref{eqn:sum-subword} if we can efficiently compute $a_{s \mid w}$.
Here we present an algorithm that computes $a_{s \mid w}$ for all $s \subword w$ in $O(\abs{w}^2)$ time.

The specific structure of the subword transition graph means that edges only go from left to right. Thus, we can split every path going through $e$ into three parts: edges left to $e$, $e$ itself and edges right to $e$. 
In terms of subwords, that is, for $s = w[i:j]$, $l=\abs{w}$, 
each segmentation $g$ that contains $s$ can be divided into three parts: segmentation $g_{w[1:i-1]}$ over $w[1:i-1]$, subword $s$ itself, and segmentation $g_{w[j+1:l]}$ over $w[j+1:l]$.
Based on this, we can rewrite \eqnref{eqn:subword-weight} as
\begin{align}
  a_{s \mid w} 
  \propto & 
    \sum_{\substack{g \in Seg_w \\ g \ni s}}
        p_s  
        {\prod_{s' \in g_{w[1:i-1]}} p_{s'}} 
        {\prod_{s' \in g_{w[j+1:l]}} p_{s'}} \\
  = & \ \ 
  p_s
  {b_{1, i-1}} 
  {b_{j+1, l}} ,
\end{align}
where  $b_{i',j'} = \sum_{g' \in Seg_{w[i':j']}} \prod_{s' \in g'} p_{s'}$.

Now we can efficiently compute $a_{s \mid w}$ if we can efficiently compute $b_{1,i-1}$ and $b_{j+1, l}$ for all $1 \le i,j \le l$. 
Fortunately, we can do so for $b_{1,i}$ using the following recursive relation 
\begin{equation}
  \label{eqn:path-sum-induction}
  b_{1,i} = \sum_{k =0}^{i-1} b_{1, k} p_{w[k+1: i]}
\end{equation}
for $i = 1, \dots, l$ with $b_{1,0} = 1$.
Similar formulas hold for $b_{j, l}, j = 1, \dots, l$ with $b_{l+1,l}=1$.

Based on this, we devise \algref{alg:subword-weights} for computing $a_{s \mid w}$ for all $s \subword w$.
Here we take the alternative view of our model as a weighted average of all possible subwords (thus the normalization in Line 12),
and an extension to the unweighted averaging of subwords as used in \citet{zhao2018generalizing-bos}.

\begin{algorithm}[tb]
   \caption{Computing $a_{s \mid w}$.}
   \label{alg:subword-weights}
\begin{algorithmic}[1]
   \STATE {\bfseries Input:} Word $w$, $p_s$ for all $s \subword w$. $l=\abs{w}$.
   \STATE {$b_{1,0} \gets 1$}; {$b_{l+1,l} \gets 1$};
   \FOR {$i \gets 1 \dots l$ }
   \STATE {$b_{1,i} \gets \sum_{k = 0}^{i-1} p_{w[k+1:i]}\ b_{1, k}$}
   \STATE $b_{l-i+1,l} \gets \sum_{k = l-i+1}^{l}  p_{w[l-i+1:k]}\ b_{k+1, l}$
   \ENDFOR
   \STATE {$\tilde{a}_{s \mid w} \gets 0$ {for all} $s \subword w$}
   \FOR {$i \gets 1 \dots l$ , $j \gets i \dots l$ }
   \STATE {$\tilde{a}' \gets p_{w[i:j]}\  b_{1,i-1}\  b_{j+1,l}$}
   \STATE {$\tilde{a}_{w[i:j] \mid w} \gets \tilde{a}_{w[i:j] \mid w} + \tilde{a}'$}
   \ENDFOR
   \STATE {$a_{s \mid w} \gets \tilde{a}_{s \mid w} / \sum_{s' \subword w} \tilde{a}_{s' \mid w} $ for all $s \subword w$}
   \RETURN {$a_{\cdot \mid w}$}
\end{algorithmic}
\end{algorithm}

\paragraph{Time complexity}
As we only access each subword once in each for-statement, the number of multiplications and additions involved is bounded by the number of subword locations of $w$.
Each of Line 4 and Line 5 take $i$ multiplications and $i-1$ additions respectively. So Line 3 to Line 6 in total takes $2l^2$ computations. Line 8 to Line 11 takes $\frac{3l(l+1)}{2}$ computations. Thus, the time complexity of \algref{alg:subword-weights} is $O(l^2)$.
Given a word of length $20$, $O(l^2)$ ($20^2 = 400$) is much better than enumerating all $O(2^l)$  ($2^{20} = 1,048,576$) segmentations.

Using the setting in \secref{sec:word-sim}, \pbos only takes 30\% more time (590 $\mu$s vs 454 $\mu$s) in average than \bos (by disabling $a_{s \mid w}$ computation) to compose a 300-dimensional word embedding vector.

\section{Experiments}
\label{sec:experiments}
We design experiments to answer two questions:
Do the segmentation likelihood and subword weights computed by \pbos align with their meaningfulness?
Are the word embedding vectors generated by \pbos of good quality?

For the former, we inspect segmentation results and subword weights (\secref{sec:seg-exp}), 
and see how good they are at predicting word affixes (\secref{sec:seg-affix-pred}).
For the latter, we evaluate the word embeddings composed by \pbos at word similarity task (\secref{sec:word-sim}) and part-of-speech (POS) tagging task (\secref{sec:pos-tagging}).

Due to the page limit, we only report the most relevant settings and results in this section. Other details, including hardware, running time and detailed list of hyperparameters, can be found in Appendix~\ref{sec:exp-details}.

\subsection{Subword Segmentation}
\label{sec:seg-exp}
In this subsection, we provide anecdotal evidence that \pbos is able to assign meaningful  segmentation likelihood and subword weights.

\begin{table*}
  \small
  \centering
  \begin{tabular}{ p{0.085\textwidth} | p{0.47\textwidth} | p{0.37\textwidth} }
    \hline\hline
      Word $w$ & Top segmentation $g$ (and their $p_{g \mid w}$) & Top subword $s$ (and their $a_{s \mid w}$) \\
    \hline
    
higher
& higher (0.924), high/er (0.030), highe/r (0.027), h/igher (0.007), hig/her (0.004).%
& higher (0.852), high (0.031), er (0.029), r (0.029), highe (0.025).%
\\

farmland
& farmland (0.971), farmlan/d (0.010), farm/land (0.006), f/armland (0.005).%
& farmland (0.941), d (0.010), farmlan (0.009), farm (0.008), land (0.007).%
\\

penpineap- plepie
& pen/pineapple/pie (0.359), pen/pineapple/pi/e (0.157), pen/pineapple/p/ie (0.101).%
& pineapple (0.238), pen (0.186), pie (0.131), p (0.101), e (0.099).%
\\

paradichlo- robenzene
& para/dichlorobenzene (0.611), par/a/dichlorobenzene (0.110), paradi/chlorobenzene (0.083).%
& dichlorobenzene (0.344), para (0.283), a (0.061), par (0.054), ichlorobenzene
(0.042).%
\\

    \hline\hline
  \end{tabular}
  \caption{\small
    Top segmentations and subword weights by PBoS for some example words
  }
  \label{tab:morph-seg}
\end{table*}

\tabref{tab:morph-seg} shows top subword segmentations and subsequent top subwords calculated by \pbos for some example word, ranked by their likelihood and weights respectively.
The calculation is based on the word frequency derived from the Google Web Trillion Word Corpus~\footnote{{\url{https://www.kaggle.com/rtatman/english-word-frequency}}}.
We use the same list for word probability $p_w$ throughout our experiments if not otherwise mentioned.
All other settings are the same as described for \pbos in \secref{sec:word-sim}.

We can see the segmentation likelihood and subword weight favors the whole words %
as subword segments if the word appears in the word list, e.g. ``higher'', ``farmland''.
This allows the model to closely mimic the word embeddings for frequent words that are probably part of the target vectors.

Second to the whole-word segmentation, or when the word is rare, e.g. ``penpineanpplepie'', ``paradichlorobenzene'',
we see that \pbos gives higher likelihood to meaningful segmentations such as ``high/er'', ``farm/land'', ``pen/pineapple/pie'' and ``para/dichlorobenzene''%
against other possible segmentations.~\footnote{A slight exception is ``farmlan/d'', probably because ``-d'' is a frequent suffix.}
Subsequently, respective subword segments get higher weights among all possible subwords for the word, often by a good amount.
This behavior would help \pbos to focus on meaningful subwords when composing word embedding.
The fact that this can be achieved without any explicit source of morphological knowledge is itself interesting.

\subsection{Affix Prediction}
\label{sec:seg-affix-pred}
We quantitatively evaluate the quality of subword segmentations and subsequent subword weights by testing if our \pbos model is able to discover the most eminent word affixes. Note this has nothing to do with embeddings, so no training is involved in this experiment.

The affix prediction task is to predict the most eminent affix for a given word.
For example, ``-able'' for ``replaceable'' and ``re-'' for ``rename''.

\paragraph{Models}
We get affix prediction from our \pbos by taking the top-ranked subword that is one of the possible affixes. 
To show our advantage, we compare it with a \bos-style baseline affix predictor.
Because \bos gives same weight to all subwords in a given word, we randomly choose one of the possible affixes that appear as subword of the word.

\paragraph{Benchmark}
We use the derivational morphology dataset~\footnote{\url{http://marcobaroni.org/PublicData/affix_complete_set.txt.gz}} from \citet{lazaridou2013compositional}.
The dataset contains 7449 English words in total along with their most eminent affixes. 
Because no training is needed in this experiment, we use all the words for evaluation. 
To make the task more challenging, we drop trivial instances where there is only one possible affix appears as a subword in the given word. 
For example, ``rename'' is dropped because only prefix ``re-'' is present; on the other hand, ``replaceable'' is kept because both ``re-'' and ``-able'' are present.
Besides excluding the trivial cases described above, we also exclude instances labeled with suffix ``-y'', because it is always included by ``-ly'' and ``-ity''.
Altogether, we acquire 3546 words with 17 possible affixes for this evaluation.

\paragraph{Results}
Affix prediction results in terms of macro precision, recall, and F1 score are shown in \tabref{tab:seg-affix-pred}.
We can see a definite advantage of \pbos at predicting most word affixes, where all the metrics boost about 0.4 and F1 almost doubles compared to \bos, providing evidence that \pbos is able to assign meaningful subword weights.

\begin{table}
  \small
  \centering
  \begin{tabular}{ l | r r r }
    \hline\hline
    Model  & Precision & Recall & F1 \\
    \hline
    \bos    &  0.493 & 0.465 & 0.425 \\ 
    \pbos   &  0.861 & 0.874 & 0.829 \\
    \hline\hline
  \end{tabular}
  \caption{\small
    Affix prediction results based on subword weights. 
    All metrics are macro.
  }
  \label{tab:seg-affix-pred}
\end{table}

\subsection{Word Similarity}
\label{sec:word-sim}
Given that \pbos is able to produce sensible  segmentation likelihood and subword weights, we now turn our focus onto the quality of the generated word embeddings.
In this section, we evaluate the word vectors' ability to capture word senses using the intrinsic task of word similarity.

Word similarity aims to test how well word embeddings capture words' semantic similarity.
The task is given as pairs of words, along with their similarity scores labeled by language speakers.
Given a set of word embeddings, we compute the similarity scores induced by the cosine distance between the embedding vectors of each pair of words.
The performance is then measured in Spearman’s correlation $\rho$ for all pairs.

\paragraph{Benchmarks}
We use WordSim353 (WS) from \citet{finkelstein2001placing-ws353} which mainly consists of common words.
To better access models' ability to generalize word embeddings towards OOV words, we include the rare word datasets RareWord (RW) from \citet{luong2013better-rw} and the newer Card-660 (Card) from \citet{pilehvar2018card}.

\paragraph{Model Setup}
\pbos composes word embeddings out of subword vectors exactly as described in \secref{sec:algorithm}.
Unlike some of previous models, we do not add special characters to indicate word boundaries
and do not set any constraint on subword lengths.
\pbos is trained 50 epochs using vanilla SGD with initial learning rate 1 and inverse square root decay. 

For baselines, we compare against the bag-of-subword model (\bos) from \citet{zhao2018generalizing-bos}, and the best attention-based model (KVQ-FH) from \citet{sasaki2019subword}. 
For \bos, we use our implementation by disabling subword weight computation.
For KVQ-FH, we use the implementation given in the paper.
All the hyperparameters are set the same as described in the original papers.
We choose to not include the character-RNN model (MIMICK) from \citet{pinter2017mimicking}, as it has been shown clearly outperformed by the two.

\paragraph{Target Vectors}
We train all the subword models over English Polyglot vectors~\footnote{\url{https://polyglot.readthedocs.io/en/latest/Download.html}} and English Google News vectors~\footnote{\url{https://code.google.com/archive/p/word2vec/}}.
Following the protocol of \citet{zhao2018generalizing-bos} and \citet{sasaki2019subword}, we clean and filter the words in Google vectors.
Dimension of word vectors, number of words in target vectors are summarized in \tabref{tab:target-stats}, along with their word similarity scores and OOV rate over the benchmarks.
As we can see, both pre-trained embeddings yield decent correlations with human-labeled word similarity. 
However, the scores drop significantly as the OOV rate goes up.
Polyglot vectors yield lower scores probably due to their smaller dimension and smaller token coverage.

\begin{table}
  \small
  \centering
  \begin{tabular}{ c | r r r }
    \hline\hline
   & WS & RW & Card \\
    \hline
    \multicolumn{4}{l}{Polyglot: 100k tokens $\times$ 64 dim} \\
    \hline
     IV pairs & 45  & 41   & 10 \\
     All pairs & 36  & 10   & 5 \\
     OOV \% & 5\% & 58\% & 90\% \\
    \hline
    \multicolumn{4}{l}{Google: 160k tokens $\times$ 300 dim} \\
    \hline
     IV pairs & 69 &  53   & 34 \\
     All pairs & 68  & 45   & 10 \\
     OOV \% & 1\% & 11\% & 79\% \\
    \hline\hline
  \end{tabular}
  \caption{\small
    Target vectors statistics and word similarity performance measured in Spearman's $\rho \times 100$.
  }
  \label{tab:target-stats}
\end{table}

\begin{table}
  \small
  \centering
  \begin{tabular}{l | r | r r r }
    \hline\hline
    Model        & \# Param  & WS & RW & Card  \\
    \hline
    \multicolumn{5}{l}{Target: Polyglot} \\
    \hline
    \bos   & 29.8M  & 34           & \textbf{34} & 6          \\
    KVQ-FH & 7.8M   & 31           & 32          & 12          \\
    \pbos  & 37.8M  & \textbf{41}  & 25          & \textbf{15} \\
    \hline
    \multicolumn{5}{l}{Target: Google News} \\
    \hline
    \bos   & 162.7M & 61             & 48          & 11          \\
    KVQ-FH & 36.2M  & 64        & \textbf{49}         & 21 \\
    \pbos  & 315.7M & \textbf{68}    & \textbf{49} & \textbf{25}          \\
    \hline\hline
  \end{tabular}
  \caption{\small
    Word similarity performance of subword-level models measured in Spearman's $\rho \times 100$.
  }
  \label{tab:word-sim-results}
\end{table}

\paragraph{Results}
Word similarity results of the three subword-level models are summarized in \tabref{tab:word-sim-results}.~\footnote{We regard training and prediction time as less of a concern here as all the three models are able to finish a training epoch in under a minute. Details and discussions can be found in Appendix~\ref{sec:exp-details-ws}.}
\pbos achieves scores better than or at least comparable to \bos and KVQ-FH in all but one of the six combinations of target vectors and word similarity benchmarks. 
Viewed as an extension to \bos, \pbos is in majority cases better than \bos, often by a good margin, suggesting the effectiveness of the subword weighting scheme.
Compared to KVQ-FH, \pbos can often match and sometimes surpass it even though \pbos is a much simpler model with better explainability.
Compared to the scores by using just the target embeddings (\tabref{tab:target-stats}, All pairs), \pbos is the only model that demonstrates improvement across \emph{all} cases.

The only case where \pbos is not doing well is with Polyglot vectors and RW benchmark.
After many manual inspections, we conjecture that it may be related to the vector norm.
Sometimes the vector of a relevant subword can be of a small norm, prone to be overwhelmed by less relevant subword vectors.
To counter this, we tried to normalize subword vectors before summing them up into a word vector (\pbosn).
\pbosn showed good improvement for the Polyglot RW case (25 to 32), matching the performance of the other two.

One may argue that \pbos has an advantage for using the most number of parameters.
However, this is largely because we do not constrain the length of subwords as in BoS or use hashing as in KVQ-FH.
In fact, restricting subword length and using hashing \emph{helped} them for the word similarity task.
We found that \pbos is insensitive to subword length constraints and decide to keep the setting simple.
Despite being an interesting direction, we decide to not involve hashing in this work to focus on the effect of our unique weighting scheme.

\paragraph{FaxtText Comparison} 
Albeit targeted for a different task (\emph{training} word embedding) which have access to contextual information, the popular fastText~\citep{bojanowski2017enriching-fasttext} also uses a subword-level model. 
We train fastText~\footnote{\url{https://github.com/facebookresearch/fastText/}} over the same English corpus on which the Polyglot target vectors are trained, in order to understand the quantitative impact of contextual information. 
To ensure a fair comparison, we restrict the vocabulary sizes and embedding dimensions to match those of Polyglot vectors. 
The word similarity scores we get for the trained fastText model are 65/40/14 for WS/RW/Card. 
We note the great gain for WS and RW, suggesting the helpfulness of contextual information in learning and generalizing word embeddings in the setting of small to moderate OOV rates. 
Surprisingly, we find that for the case of extremely high OOV rate (Card), \pbos slightly surpasses fastText, suggesting \pbos' effectiveness in generalizing embeddings to OOV words even without any help from contexts.

\paragraph{Multilingual Results}
To evaluate and compare the effectiveness of \pbos across languages, we further train the models targeting multilingual Wikipedia2Vec vectors~\citep{yamada2020wikipedia2vec} and evaluate them on multilingual WordSim353 and SemLex999 from \citet{leviant2015separated} which are available in English, German, Italian and Russian.
To better access the models' ability to \emph{generalize}, we only take the top 10k words from the target vectors for training, which yields decent OOV rates, ranging from 23\% to 84\%.
Detailed results can be found in Appendix \secref{sec:exp-details-multilingual-ws}.
In summary, we find 1) that \pbos surpasses KVQ-FH for English and German and is comparable to KVQ-FH for Italian; 2) that \pbos and KVQ-FH surpasses \bos for English, German and Italian; and 3) no definitive trend among the three models for Russian.

\subsection{POS Tagging}
\label{sec:pos-tagging}
We further assess the quality of generated word embedding via the extrinsic task of POS tagging.
The task is to categorize each word in a given context into a particular part of speech, e.g. noun, verb, and adjective.

\paragraph{POS Tagging Model}
We follow the evaluation protocol for sequential labeling used by \citet{kiros2015skip} and \citet{li2017investigating}, and use logistic regression classifier~\footnote{\url{https://scikit-learn.org/0.19/modules/generated/sklearn.linear_model.LogisticRegression.html}} as the model for POS tagging.
When predicting the tag for the $i$-th word $w_i$ in a sentence, 
the input to the classifier is the concatenation of the vectors $\vec{w}_{i-2}, \vec{w}_{i-1}, \vec{w}_{i}, \vec{w}_{i+1}, \vec{w}_{i+2}$ for the word itself and the words in its context.
This setup allows a more direct evaluation of the quality of word vectors themselves, and thus gives better discriminative power.~\footnote{
As a side note, in our early trials, we tried to evaluate using an LSTM model following \citet{pinter2017mimicking} and \citet{zhao2018generalizing-bos}, but found the numbers rather similar across embedding models.
One possible explanation is that LSTMs are so good at picking up contextual features that the impact of mild deviations of a single word vector is marginal.
}

\paragraph{Dataset}
We train and evaluate the performance of generated word embeddings over 23 languages at the intersection of the Polyglot~\citep{al-rfou2013polyglot} pre-trained embedding vectors~\footnote{\url{https://polyglot.readthedocs.io/}} and the Universal Dependency~(UD, v1.4~\footnote{\url{https://universaldependencies.org/}}) dataset.
Polyglot vectors contain 64-dimensional vectors over an 100k vocabulary for each language and are used as target vectors for each of the subword-level embedding models in this experiment.
For \pbos, we use the Polyglot word counts for each language as the base for subword segmentation and subword weights calculation.
UD is used as the POS tagging dataset to train and test the POS tagging model.
We use the default partition of training and testing set.
Statistics vary from language to language. See Appendix~\ref{sec:exp-details-pos} for more details.

\begin{table}[!t]
  \small
  \centering
  \begin{tabular}{ l | c c c }
    \hline\hline
    Language  & KVQ-FH & BoS   & \pbos  \\
    \hline

Arabic     & 0.813 & 0.754          & \textbf{0.905}(+0.092)  \\
Basque     & 0.749 & 0.829          & \textbf{0.866}(+0.037)  \\
Bulgarian  & 0.777 & 0.793          & \textbf{0.929}(+0.136)  \\
Chinese    & 0.633 & 0.330          & \textbf{0.833}(+0.200)  \\
Czech      & 0.799 & 0.823          & \textbf{0.917}(+0.094)  \\
Danish     & 0.801 & 0.757          & \textbf{0.904}(+0.103)  \\
English    & 0.770 & 0.770          & \textbf{0.896}(+0.126)  \\
Greek      & 0.866 & 0.888          & \textbf{0.941}(+0.053)  \\
Hebrew     & 0.775 & 0.703          & \textbf{0.915}(+0.140)  \\
Hindi      & 0.811 & 0.800          & \textbf{0.901}(+0.090)  \\
Hungarian  & 0.777 & 0.794          & \textbf{0.893}(+0.099)  \\
Indonesian & 0.776 & 0.828          & \textbf{0.899}(+0.071)  \\
Italian    & 0.794 & 0.787          & \textbf{0.930}(+0.135)  \\
Kazakh     & 0.623 & 0.753          & \textbf{0.815}(+0.062)  \\
Latvian    & 0.722 & 0.756          & \textbf{0.848}(+0.092)  \\
Persian    & 0.869 & 0.782          & \textbf{0.924}(+0.056)  \\
Romanian   & 0.774 & 0.755          & \textbf{0.898}(+0.123)  \\
Russian    & 0.775 & 0.838          & \textbf{0.911}(+0.073)  \\
Spanish    & 0.818 & 0.769          & \textbf{0.920}(+0.102)  \\
Swedish    & 0.826 & 0.840          & \textbf{0.920}(+0.080)  \\
Tamil      & 0.702 & \textbf{0.758} & 0.755(-0.003)  \\
Turkish    & 0.760 & 0.777          & \textbf{0.837}(+0.060)  \\
Vietnamese & 0.663 & 0.712          & \textbf{0.832}(+0.121)  \\
    \hline\hline
  \end{tabular}
    
  \caption{\small
        POS tagging accuracy over 23 languages.
        In parentheses are the gains to the best of \bos and KVQ-FH.
  }
  \label{tab:pos-tagging-results}
\end{table}
\paragraph{Results}
\tabref{tab:pos-tagging-results} shows the POS tagging accuracy over the 23 languages that appear in both Polyglot and UD.
All the subword-level embedding models follow the same hyperparameters as in \secref{sec:word-sim}.
Following \citet{sasaki2019subword}, we tune the regularization term of the logistic regression model when evaluating KVQ-FH.
Even with that, \pbos is able to achieve the best POS tagging accuracy in all but one language regardless of morphological types, OOV rates, and the number of training instances (Appendix~\tabref{tab:pos-tagging-results-full}).
Particularly, \pbos improvement accuracy by greater than 0.1 for 9 languages.
For the one language (Tamil) where \pbos is not the most accurate, the difference to the best is small (0.003).
KVQ-FH gives no significantly more accurate predictions than \bos despite it is more complex and is the only one tuned with hyperparameters. 

Overall, \tabref{tab:pos-tagging-results} shows that the word embeddings composed by our \pbos is effective at predicting POS tags for a wide range of languages.

\section{Related Work}
Popular word embedding methods, such as word2vec~\citep{mikolov2013distributed-w2v}, GloVe~\citep{pennington2014glove}, often assume finite-size vocabularies,
giving rise to the problem of OOV words.

FastText~\citep{bojanowski2017enriching-fasttext, joulin2017bag} attempted to alleviate the problem using subword-level model,
and was followed by interests of using subword information to improve word embedding~\citep{wieting2016charagram, cao2017improving, li2017investigating, athiwaratkun2018probabilistic, li2018subword, salle2018incorporating, xu2019treat, zhu2019systematic}. 
Among them are Charagram by \citet{wieting2016charagram} which, albeit trained on specific downstream tasks, is similar to \bos followed by a non-linear activation, and the systematic evaluation by \citet{zhu2019systematic} over various choices of word composition functions and subword segmentation methods.
However, all works above either pay little attention to the interaction among subwords inside a given word, or treat subword segmentation and composing word representation as separate problems.

Another interesting thread of works~\citep{oshikiri2017segmentation, kim2018word-ngram-embedding, kim2019segmentation-ngram-embedding} attempted to model language solely at the subword level and learn subword embeddings directly from text,
providing evidence to the power of subword-level models,
especially as the notion of word is thought doubtful by some linguistics~\citep{haspelmath2011indeterminacy}.

Besides the recent interest in subwords, there have been long efforts of using morphology to improve word embedding ~\citep{luong2013better-rw, cotterell2015morphological, cui2015knet, soricut2015unsupervised, bhatia2016morphological, cao2016joint, xu2018incorporating, ustun2018characters, edmiston2018compositional, chaudhary2018adapting, park2018grapheme}.
However, most of them require an external oracle, such as Morfessor~\citep{creutz2002unsupervised, virpioja2013morfessor}, for the morphological segmentations of input words, limiting their power to the quality and availability of such segmenters.
The only exception is the character LSTM model by \citet{cao2016joint}, which has shown some ability to recover the morphological boundary as a byproduct of learning word embedding.

The most related works in generalizing pre-trained word embeddings have been discussed in \secref{sec:introduction} and compared throughout the paper.

\section{Conclusion and Future Work}
We propose \pbos model for generalizing pre-trained word embeddings without contextual information.
\pbos simultaneously considers all possible subword segmentations of a word and derives meaningful subword weights that lead to better composed word embeddings.
Experiments on segmentation results, affix prediction, word similarity, and POS tagging over 23 languages support the claim.

In the future, it would be interesting to see if \pbos can also help with the task of \emph{learning} word embedding, and how hashing would impact the quality of composed embedding while facilitating a more compact model.

\section*{Acknowledgments}

The authors would like to thank anonymous reviewers of EMNLP for their comments.
ZJ would like to thank Xuezhou Zhang, Sidharth Mudgal, Matt Du and Harit Vishwakarma for their helpful discussions.

\bibliography{main}

\begin{thebibliography}{45}
\expandafter\ifx\csname natexlab\endcsname\relax\def\natexlab#1{#1}\fi

\bibitem[{Al-Rfou{'} et~al.(2013)Al-Rfou{'}, Perozzi, and
  Skiena}]{al-rfou2013polyglot}
Rami Al-Rfou{'}, Bryan Perozzi, and Steven Skiena. 2013.
\newblock \href {https://www.aclweb.org/anthology/W13-3520} {{P}olyglot:
  Distributed word representations for multilingual {NLP}}.
\newblock In \emph{Proceedings of the Seventeenth Conference on Computational
  Natural Language Learning}, pages 183--192, Sofia, Bulgaria. Association for
  Computational Linguistics.

\bibitem[{Athiwaratkun et~al.(2018)Athiwaratkun, Wilson, and
  Anandkumar}]{athiwaratkun2018probabilistic}
Ben Athiwaratkun, Andrew Wilson, and Anima Anandkumar. 2018.
\newblock \href {https://doi.org/10.18653/v1/P18-1001} {Probabilistic
  {F}ast{T}ext for multi-sense word embeddings}.
\newblock In \emph{Proceedings of the 56th Annual Meeting of the Association
  for Computational Linguistics (Volume 1: Long Papers)}, pages 1--11,
  Melbourne, Australia. Association for Computational Linguistics.

\bibitem[{Bhatia et~al.(2016)Bhatia, Guthrie, and
  Eisenstein}]{bhatia2016morphological}
Parminder Bhatia, Robert Guthrie, and Jacob Eisenstein. 2016.
\newblock \href {https://doi.org/10.18653/v1/D16-1047} {Morphological priors
  for probabilistic neural word embeddings}.
\newblock In \emph{Proceedings of the 2016 Conference on Empirical Methods in
  Natural Language Processing}, pages 490--500, Austin, Texas. Association for
  Computational Linguistics.

\bibitem[{Bojanowski et~al.(2017)Bojanowski, Grave, Joulin, and
  Mikolov}]{bojanowski2017enriching-fasttext}
Piotr Bojanowski, Edouard Grave, Armand Joulin, and Tomas Mikolov. 2017.
\newblock \href {https://doi.org/10.1162/tacl_a_00051} {Enriching word vectors
  with subword information}.
\newblock \emph{Transactions of the Association for Computational Linguistics},
  5:135--146.

\bibitem[{Cao and Rei(2016)}]{cao2016joint}
Kris Cao and Marek Rei. 2016.
\newblock \href {https://doi.org/10.18653/v1/W16-1603} {A joint model for word
  embedding and word morphology}.
\newblock In \emph{Proceedings of the 1st Workshop on Representation Learning
  for {NLP}}, pages 18--26, Berlin, Germany. Association for Computational
  Linguistics.

\bibitem[{Cao and Lu(2017)}]{cao2017improving}
Shaosheng Cao and Wei Lu. 2017.
\newblock \href
  {https://www.aaai.org/ocs/index.php/AAAI/AAAI17/paper/view/14724} {Improving
  word embeddings with convolutional feature learning and subword information}.
\newblock In \emph{Proceedings of the AAAI Conference on Artificial
  Intelligence}.

\bibitem[{Chaudhary et~al.(2018)Chaudhary, Zhou, Levin, Neubig, Mortensen, and
  Carbonell}]{chaudhary2018adapting}
Aditi Chaudhary, Chunting Zhou, Lori Levin, Graham Neubig, David~R. Mortensen,
  and Jaime Carbonell. 2018.
\newblock \href {https://doi.org/10.18653/v1/D18-1366} {Adapting word
  embeddings to new languages with morphological and phonological subword
  representations}.
\newblock In \emph{Proceedings of the 2018 Conference on Empirical Methods in
  Natural Language Processing}, pages 3285--3295, Brussels, Belgium.
  Association for Computational Linguistics.

\bibitem[{Cotterell and Sch{\"u}tze(2015)}]{cotterell2015morphological}
Ryan Cotterell and Hinrich Sch{\"u}tze. 2015.
\newblock \href {https://doi.org/10.3115/v1/N15-1140} {Morphological
  word-embeddings}.
\newblock In \emph{Proceedings of the 2015 Conference of the North {A}merican
  Chapter of the Association for Computational Linguistics: Human Language
  Technologies}, pages 1287--1292, Denver, Colorado. Association for
  Computational Linguistics.

\bibitem[{Cotterell et~al.(2016)Cotterell, Sch{\"u}tze, and
  Eisner}]{cotterell2016morphological}
Ryan Cotterell, Hinrich Sch{\"u}tze, and Jason Eisner. 2016.
\newblock \href {https://doi.org/10.18653/v1/P16-1156} {Morphological smoothing
  and extrapolation of word embeddings}.
\newblock In \emph{Proceedings of the 54th Annual Meeting of the Association
  for Computational Linguistics (Volume 1: Long Papers)}, pages 1651--1660,
  Berlin, Germany. Association for Computational Linguistics.

\bibitem[{Creutz and Lagus(2002)}]{creutz2002unsupervised}
Mathias Creutz and Krista Lagus. 2002.
\newblock \href {https://doi.org/10.3115/1118647.1118650} {Unsupervised
  discovery of morphemes}.
\newblock In \emph{Proceedings of the {ACL}-02 Workshop on Morphological and
  Phonological Learning}, pages 21--30. Association for Computational
  Linguistics.

\bibitem[{Cui et~al.(2015)Cui, Gao, Bian, Qiu, Dai, and Liu}]{cui2015knet}
Qing Cui, Bin Gao, Jiang Bian, Siyu Qiu, Hanjun Dai, and Tie-Yan Liu. 2015.
\newblock \href {https://doi.org/10.1145/2797137} {{KNET}: A general framework
  for learning word embedding using morphological knowledge}.
\newblock \emph{ACM Trans. Inf. Syst.}, 34(1).

\bibitem[{Edmiston and Stratos(2018)}]{edmiston2018compositional}
Daniel Edmiston and Karl Stratos. 2018.
\newblock \href {https://doi.org/10.18653/v1/W18-2901} {Compositional morpheme
  embeddings with affixes as functions and stems as arguments}.
\newblock In \emph{Proceedings of the Workshop on the Relevance of Linguistic
  Structure in Neural Architectures for {NLP}}, pages 1--5, Melbourne,
  Australia. Association for Computational Linguistics.

\bibitem[{Finkelstein et~al.(2001)Finkelstein, Gabrilovich, Matias, Rivlin,
  Solan, Wolfman, and Ruppin}]{finkelstein2001placing-ws353}
Lev Finkelstein, Evgeniy Gabrilovich, Yossi Matias, Ehud Rivlin, Zach Solan,
  Gadi Wolfman, and Eytan Ruppin. 2001.
\newblock \href {https://doi.org/10.1145/371920.372094} {Placing search in
  context: The concept revisited}.
\newblock In \emph{Proceedings of the 10th International Conference on World
  Wide Web}, WWW ’01, page 406–414, New York, NY, USA. Association for
  Computing Machinery.

\bibitem[{Haspelmath(2011)}]{haspelmath2011indeterminacy}
Martin Haspelmath. 2011.
\newblock \href {https://doi.org/10.1515/flin.2011.002} {The indeterminacy of
  word segmentation and the nature of morphology and syntax}.
\newblock \emph{Folia linguistica}, 45(1):31--80.

\bibitem[{Joulin et~al.(2017)Joulin, Grave, Bojanowski, and
  Mikolov}]{joulin2017bag}
Armand Joulin, Edouard Grave, Piotr Bojanowski, and Tomas Mikolov. 2017.
\newblock \href {https://www.aclweb.org/anthology/E17-2068} {Bag of tricks for
  efficient text classification}.
\newblock In \emph{Proceedings of the 15th Conference of the {E}uropean Chapter
  of the Association for Computational Linguistics: Volume 2, Short Papers},
  pages 427--431, Valencia, Spain. Association for Computational Linguistics.

\bibitem[{Khodak et~al.(2018)Khodak, Saunshi, Liang, Ma, Stewart, and
  Arora}]{khodak2018carte}
Mikhail Khodak, Nikunj Saunshi, Yingyu Liang, Tengyu Ma, Brandon Stewart, and
  Sanjeev Arora. 2018.
\newblock \href {https://doi.org/10.18653/v1/P18-1002} {A la carte embedding:
  Cheap but effective induction of semantic feature vectors}.
\newblock In \emph{Proceedings of the 56th Annual Meeting of the Association
  for Computational Linguistics (Volume 1: Long Papers)}, pages 12--22,
  Melbourne, Australia. Association for Computational Linguistics.

\bibitem[{Kim et~al.(2018{\natexlab{a}})Kim, Fukui, and
  Shimodaira}]{kim2018word-ngram-embedding}
Geewook Kim, Kazuki Fukui, and Hidetoshi Shimodaira. 2018{\natexlab{a}}.
\newblock \href {https://doi.org/10.18653/v1/W18-6120} {Word-like character
  n-gram embedding}.
\newblock In \emph{Proceedings of the 2018 {EMNLP} Workshop W-{NUT}: The 4th
  Workshop on Noisy User-generated Text}, pages 148--152, Brussels, Belgium.
  Association for Computational Linguistics.

\bibitem[{Kim et~al.(2019)Kim, Fukui, and
  Shimodaira}]{kim2019segmentation-ngram-embedding}
Geewook Kim, Kazuki Fukui, and Hidetoshi Shimodaira. 2019.
\newblock \href {https://doi.org/10.18653/v1/N19-1324} {Segmentation-free
  compositional $n$-gram embedding}.
\newblock In \emph{Proceedings of the 2019 Conference of the North {A}merican
  Chapter of the Association for Computational Linguistics: Human Language
  Technologies, Volume 1 (Long and Short Papers)}, pages 3207--3215,
  Minneapolis, Minnesota. Association for Computational Linguistics.

\bibitem[{Kim et~al.(2018{\natexlab{b}})Kim, Kim, Lee, and
  Lee}]{kim2018learning-cnn}
Yeachan Kim, Kang-Min Kim, Ji-Min Lee, and SangKeun Lee. 2018{\natexlab{b}}.
\newblock \href {https://www.aclweb.org/anthology/C18-1216} {Learning to
  generate word representations using subword information}.
\newblock In \emph{Proceedings of the 27th International Conference on
  Computational Linguistics}, pages 2551--2561, Santa Fe, New Mexico, USA.
  Association for Computational Linguistics.

\bibitem[{Kiros et~al.(2015)Kiros, Zhu, Salakhutdinov, Zemel, Urtasun,
  Torralba, and Fidler}]{kiros2015skip}
Ryan Kiros, Yukun Zhu, Russ~R Salakhutdinov, Richard Zemel, Raquel Urtasun,
  Antonio Torralba, and Sanja Fidler. 2015.
\newblock \href {http://papers.nips.cc/paper/5950-skip-thought-vectors.pdf}
  {Skip-thought vectors}.
\newblock In C.~Cortes, N.~D. Lawrence, D.~D. Lee, M.~Sugiyama, and R.~Garnett,
  editors, \emph{Advances in Neural Information Processing Systems 28}, pages
  3294--3302. Curran Associates, Inc.

\bibitem[{Lazaridou et~al.(2013)Lazaridou, Marelli, Zamparelli, and
  Baroni}]{lazaridou2013compositional}
Angeliki Lazaridou, Marco Marelli, Roberto Zamparelli, and Marco Baroni. 2013.
\newblock \href {https://www.aclweb.org/anthology/P13-1149} {Compositional-ly
  derived representations of morphologically complex words in distributional
  semantics}.
\newblock In \emph{Proceedings of the 51st Annual Meeting of the Association
  for Computational Linguistics (Volume 1: Long Papers)}, pages 1517--1526,
  Sofia, Bulgaria. Association for Computational Linguistics.

\bibitem[{Leviant and Reichart(2015)}]{leviant2015separated}
Ira Leviant and Roi Reichart. 2015.
\newblock \href {http://arxiv.org/abs/1508.00106} {Separated by an un-common
  language: Towards judgment language informed vector space modeling}.

\bibitem[{Li et~al.(2018)Li, Drozd, Liu, and Du}]{li2018subword}
Bofang Li, Aleksandr Drozd, Tao Liu, and Xiaoyong Du. 2018.
\newblock \href {https://doi.org/10.18653/v1/W18-1205} {Subword-level
  composition functions for learning word embeddings}.
\newblock In \emph{Proceedings of the Second Workshop on Subword/Character
  {LE}vel Models}, pages 38--48, New Orleans. Association for Computational
  Linguistics.

\bibitem[{Li et~al.(2017)Li, Liu, Zhao, Tang, Drozd, Rogers, and
  Du}]{li2017investigating}
Bofang Li, Tao Liu, Zhe Zhao, Buzhou Tang, Aleksandr Drozd, Anna Rogers, and
  Xiaoyong Du. 2017.
\newblock \href {https://doi.org/10.18653/v1/D17-1257} {Investigating different
  syntactic context types and context representations for learning word
  embeddings}.
\newblock In \emph{Proceedings of the 2017 Conference on Empirical Methods in
  Natural Language Processing}, pages 2421--2431, Copenhagen, Denmark.
  Association for Computational Linguistics.

\bibitem[{Luong et~al.(2013)Luong, Socher, and Manning}]{luong2013better-rw}
Thang Luong, Richard Socher, and Christopher Manning. 2013.
\newblock \href {https://www.aclweb.org/anthology/W13-3512} {Better word
  representations with recursive neural networks for morphology}.
\newblock In \emph{Proceedings of the Seventeenth Conference on Computational
  Natural Language Learning}, pages 104--113, Sofia, Bulgaria. Association for
  Computational Linguistics.

\bibitem[{Mikolov et~al.(2013)Mikolov, Sutskever, Chen, Corrado, and
  Dean}]{mikolov2013distributed-w2v}
Tomas Mikolov, Ilya Sutskever, Kai Chen, Greg~S Corrado, and Jeff Dean. 2013.
\newblock \href
  {http://papers.nips.cc/paper/5021-distributed-representations-of-words-and-phrases-and-their-compositionality.pdf}
  {Distributed representations of words and phrases and their
  compositionality}.
\newblock In C.~J.~C. Burges, L.~Bottou, M.~Welling, Z.~Ghahramani, and K.~Q.
  Weinberger, editors, \emph{Advances in Neural Information Processing Systems
  26}, pages 3111--3119. Curran Associates, Inc.

\bibitem[{Oshikiri(2017)}]{oshikiri2017segmentation}
Takamasa Oshikiri. 2017.
\newblock \href {https://doi.org/10.18653/v1/D17-1080} {Segmentation-free word
  embedding for unsegmented languages}.
\newblock In \emph{Proceedings of the 2017 Conference on Empirical Methods in
  Natural Language Processing}, pages 767--772, Copenhagen, Denmark.
  Association for Computational Linguistics.

\bibitem[{Park and Shin(2018)}]{park2018grapheme}
Suzi Park and Hyopil Shin. 2018.
\newblock \href {https://www.aclweb.org/anthology/L18-1471} {Grapheme-level
  awareness in word embeddings for morphologically rich languages}.
\newblock In \emph{Proceedings of the Eleventh International Conference on
  Language Resources and Evaluation ({LREC} 2018)}, Miyazaki, Japan. European
  Language Resources Association (ELRA).

\bibitem[{Pennington et~al.(2014)Pennington, Socher, and
  Manning}]{pennington2014glove}
Jeffrey Pennington, Richard Socher, and Christopher Manning. 2014.
\newblock \href {https://doi.org/10.3115/v1/D14-1162} {{G}love: Global vectors
  for word representation}.
\newblock In \emph{Proceedings of the 2014 Conference on Empirical Methods in
  Natural Language Processing ({EMNLP})}, pages 1532--1543, Doha, Qatar.
  Association for Computational Linguistics.

\bibitem[{Pilehvar et~al.(2018)Pilehvar, Kartsaklis, Prokhorov, and
  Collier}]{pilehvar2018card}
Mohammad~Taher Pilehvar, Dimitri Kartsaklis, Victor Prokhorov, and Nigel
  Collier. 2018.
\newblock \href {https://doi.org/10.18653/v1/D18-1169} {Card-660: {C}ambridge
  rare word dataset - a reliable benchmark for infrequent word representation
  models}.
\newblock In \emph{Proceedings of the 2018 Conference on Empirical Methods in
  Natural Language Processing}, pages 1391--1401, Brussels, Belgium.
  Association for Computational Linguistics.

\bibitem[{Pinter et~al.(2017)Pinter, Guthrie, and
  Eisenstein}]{pinter2017mimicking}
Yuval Pinter, Robert Guthrie, and Jacob Eisenstein. 2017.
\newblock \href {https://doi.org/10.18653/v1/D17-1010} {Mimicking word
  embeddings using subword {RNN}s}.
\newblock In \emph{Proceedings of the 2017 Conference on Empirical Methods in
  Natural Language Processing}, pages 102--112, Copenhagen, Denmark.
  Association for Computational Linguistics.

\bibitem[{Salle and Villavicencio(2018)}]{salle2018incorporating}
Alexandre Salle and Aline Villavicencio. 2018.
\newblock \href {https://doi.org/10.18653/v1/W18-1209} {Incorporating subword
  information into matrix factorization word embeddings}.
\newblock In \emph{Proceedings of the Second Workshop on Subword/Character
  {LE}vel Models}, pages 66--71, New Orleans. Association for Computational
  Linguistics.

\bibitem[{Sasaki et~al.(2019)Sasaki, Suzuki, and Inui}]{sasaki2019subword}
Shota Sasaki, Jun Suzuki, and Kentaro Inui. 2019.
\newblock \href {https://doi.org/10.18653/v1/N19-1353} {{S}ubword-based
  {C}ompact {R}econstruction of {W}ord {E}mbeddings}.
\newblock In \emph{Proceedings of the 2019 Conference of the North {A}merican
  Chapter of the Association for Computational Linguistics: Human Language
  Technologies, Volume 1 (Long and Short Papers)}, pages 3498--3508,
  Minneapolis, Minnesota. Association for Computational Linguistics.

\bibitem[{Schick and Sch{\"u}tze(2019{\natexlab{a}})}]{schick2019attentive}
Timo Schick and Hinrich Sch{\"u}tze. 2019{\natexlab{a}}.
\newblock \href {https://doi.org/10.18653/v1/N19-1048} {Attentive mimicking:
  Better word embeddings by attending to informative contexts}.
\newblock In \emph{Proceedings of the 2019 Conference of the North {A}merican
  Chapter of the Association for Computational Linguistics: Human Language
  Technologies, Volume 1 (Long and Short Papers)}, pages 489--494, Minneapolis,
  Minnesota. Association for Computational Linguistics.

\bibitem[{Schick and Sch{\"u}tze(2019{\natexlab{b}})}]{schick2019learning}
Timo Schick and Hinrich Sch{\"u}tze. 2019{\natexlab{b}}.
\newblock \href {https://doi.org/10.1609/aaai.v33i01.33016965} {Learning
  semantic representations for novel words: Leveraging both form and context}.
\newblock In \emph{Proceedings of the AAAI Conference on Artificial
  Intelligence}, volume~33, pages 6965--6973.

\bibitem[{Soricut and Och(2015)}]{soricut2015unsupervised}
Radu Soricut and Franz Och. 2015.
\newblock \href {https://doi.org/10.3115/v1/N15-1186} {Unsupervised morphology
  induction using word embeddings}.
\newblock In \emph{Proceedings of the 2015 Conference of the North {A}merican
  Chapter of the Association for Computational Linguistics: Human Language
  Technologies}, pages 1627--1637, Denver, Colorado. Association for
  Computational Linguistics.

\bibitem[{Stratos(2017)}]{stratos2017reconstruction-rnn}
Karl Stratos. 2017.
\newblock \href {https://doi.org/10.18653/v1/W17-4119} {Reconstruction of word
  embeddings from sub-word parameters}.
\newblock In \emph{Proceedings of the First Workshop on Subword and Character
  Level Models in {NLP}}, pages 130--135, Copenhagen, Denmark. Association for
  Computational Linguistics.

\bibitem[{{\"U}st{\"u}n et~al.(2018){\"U}st{\"u}n, Kurfal{\i}, and
  Can}]{ustun2018characters}
Ahmet {\"U}st{\"u}n, Murathan Kurfal{\i}, and Burcu Can. 2018.
\newblock \href {https://doi.org/10.18653/v1/W18-3019} {Characters or
  morphemes: How to represent words?}
\newblock In \emph{Proceedings of The Third Workshop on Representation Learning
  for {NLP}}, pages 144--153, Melbourne, Australia. Association for
  Computational Linguistics.

\bibitem[{Virpioja et~al.(2013)Virpioja, Smit, Grönroos, and
  Kurimo}]{virpioja2013morfessor}
Sami Virpioja, Peter Smit, Stig-Arne Grönroos, and Mikko Kurimo. 2013.
\newblock \href {http://urn.fi/URN:ISBN:978-952-60-5501-5} {Morfessor 2.0:
  Python implementation and extensions for morfessor baseline}.
\newblock D4 julkaistu kehittämis- tai tutkimusraportti tai -selvitys, Aalto
  University.

\bibitem[{Wieting et~al.(2016)Wieting, Bansal, Gimpel, and
  Livescu}]{wieting2016charagram}
John Wieting, Mohit Bansal, Kevin Gimpel, and Karen Livescu. 2016.
\newblock \href {https://doi.org/10.18653/v1/D16-1157} {{C}haragram: Embedding
  words and sentences via character n-grams}.
\newblock In \emph{Proceedings of the 2016 Conference on Empirical Methods in
  Natural Language Processing}, pages 1504--1515, Austin, Texas. Association
  for Computational Linguistics.

\bibitem[{Xu et~al.(2018)Xu, Liu, Yang, and Huang}]{xu2018incorporating}
Yang Xu, Jiawei Liu, Wei Yang, and Liusheng Huang. 2018.
\newblock \href {https://doi.org/10.18653/v1/P18-1114} {Incorporating latent
  meanings of morphological compositions to enhance word embeddings}.
\newblock In \emph{Proceedings of the 56th Annual Meeting of the Association
  for Computational Linguistics (Volume 1: Long Papers)}, pages 1232--1242,
  Melbourne, Australia. Association for Computational Linguistics.

\bibitem[{Xu et~al.(2019)Xu, Zhang, and Reitter}]{xu2019treat}
Yang Xu, Jiasheng Zhang, and David Reitter. 2019.
\newblock \href {https://doi.org/10.18653/v1/W19-4717} {Treat the word as a
  whole or look inside? subword embeddings model language change and typology}.
\newblock In \emph{Proceedings of the 1st International Workshop on
  Computational Approaches to Historical Language Change}, pages 136--145,
  Florence, Italy. Association for Computational Linguistics.

\bibitem[{Yamada et~al.(2020)Yamada, Asai, Sakuma, Shindo, Takeda, Takefuji,
  and Matsumoto}]{yamada2020wikipedia2vec}
Ikuya Yamada, Akari Asai, Jin Sakuma, Hiroyuki Shindo, Hideaki Takeda,
  Yoshiyasu Takefuji, and Yuji Matsumoto. 2020.
\newblock \href {http://arxiv.org/abs/1812.06280} {Wikipedia2vec: An efficient
  toolkit for learning and visualizing the embeddings of words and entities
  from wikipedia}.

\bibitem[{Zhao et~al.(2018)Zhao, Mudgal, and Liang}]{zhao2018generalizing-bos}
Jinman Zhao, Sidharth Mudgal, and Yingyu Liang. 2018.
\newblock \href {https://doi.org/10.18653/v1/D18-1059} {Generalizing word
  embeddings using bag of subwords}.
\newblock In \emph{Proceedings of the 2018 Conference on Empirical Methods in
  Natural Language Processing}, pages 601--606, Brussels, Belgium. Association
  for Computational Linguistics.

\bibitem[{Zhu et~al.(2019)Zhu, Vuli{\'c}, and Korhonen}]{zhu2019systematic}
Yi~Zhu, Ivan Vuli{\'c}, and Anna Korhonen. 2019.
\newblock \href {https://doi.org/10.18653/v1/N19-1097} {A systematic study of
  leveraging subword information for learning word representations}.
\newblock In \emph{Proceedings of the 2019 Conference of the North {A}merican
  Chapter of the Association for Computational Linguistics: Human Language
  Technologies, Volume 1 (Long and Short Papers)}, pages 912--932, Minneapolis,
  Minnesota. Association for Computational Linguistics.

\end{thebibliography}
\bibliographystyle{acl_natbib}

\clearpage
\newpage
\appendix

\section{Experimental Details}
\label{sec:exp-details}
Here we list the details of our experiments that are omitted in the main paper due to space constraints.

We run all our experiments on a machine with an 8-core Intel i7-6700 CPU @ 3.40GHz, 32GB Memory, and GeForce GTX 970 GPU.

\subsection{Hyperparameters}
\label{sec:hparams}
The meaning of hyperparameters shown in \tabref{tab:ws-settings}, \tabref{tab:pos-settings} and \tabref{tab:sasaki-settings} as explained as follows.

\paragraph{Subwords}
\begin{itemize}[leftmargin=*]
\setlength\itemsep{-0.5em}
    \item \code{min\_len}: The minimum length for a subword to be considered.
    \item \code{max\_len}: The maximum length for a subword to be considered.
    \item \code{word\_boundary}: Whether to add special characters to annotate word boundaries.
\end{itemize}
\paragraph{Training}
\begin{itemize}[leftmargin=*]
\setlength\itemsep{-0.5em}
    \item \code{epochs}: The number of training epochs.
    \item \code{lr}: Learning rate.
    \item \code{lr\_decay}: Whether to set learning rate to be inversely proportional to the square root of the epoch number.
    \item \code{normalize\_semb}: Whether to normalize subword embeddings before composing word embeddings.
    \item \code{prob\_eps}: Default likelihood for  unknown characters. 
\end{itemize}
\paragraph{Evaluation}
\begin{itemize}[leftmargin=*]
\setlength\itemsep{-0.5em}
    \item \code{C}: The inverse regularization term used by the logistic regression classifier.
\end{itemize}

\subsection{Word Similarity}
\label{sec:exp-details-ws}
\tabref{tab:ws-settings} and \tabref{tab:sasaki-settings} show the hyperparameter values used in the word similarity experiment (\secref{sec:word-sim}). 
We transform all words in the benchmarks into lowercase, following the convention in FastText~\citep{bojanowski2017enriching-fasttext, joulin2017bag}, \bos~\citep{zhao2018generalizing-bos}, and KVQ-FH~\citep{sasaki2019subword}. 

During the evaluation, we use 0 as the similarity score for a pair of words if we cannot get word vector for one of the words, or the magnitude of the word vector is too small.
This is especially the case when we evaluate the target vectors, where OOV rates can be significant.

\tabref{tab:word-sim-results-full} lists experimental result for word similarity in greater detail. 

Regarding the training epoch time, note that KVQ-FH uses GPU and is implemented using a deep learning library~\footnote{Chainer, \url{https://chainer.org/}} with underlying optimized C code, whereas our \pbos is implemented using pure Python and uses only single thread CPU. 
We omit the prediction time for KVQ-FH, as we found it hard to separate the actual inference time from time used for other processes such as batching and data transfer between CPU and GPU. 
However, we believe the overall trend should be similar as for the training time.

One may notice that the prediction time for \bos in \tabref{tab:word-sim-results-full} is different from what was reported at the end of \secref{sec:algorithm}.
This is largely because the \bos in \tabref{tab:word-sim-results-full} has a different (smaller) set of possible subwords to consider due to the subword length limits.
In \secref{sec:algorithm}, to fairly access the impact of subword weights computation, we ensure that \bos and \pbos work with the same set of possible subwords (that used by \pbos in \secref{sec:word-sim}), and thus observe a slight longer prediction time for \bos.

\subsection{Multilingual Word Similarity}
\label{sec:exp-details-multilingual-ws}

We use Wikipedia2Vec~\citep{yamada2020wikipedia2vec} as target vectors, and keep the most frequent 10k words to get decent OOV rates. 
The OOV rates and word similarity scores can be found in \tabref{tab:word-sim-multilingual-target-stats}.

We do not clean or filter words as we did for the English word similarity, because we found it difficult to have a consistent way of pre-processing words across languages.
For \pbos, we use the word frequencies from Polyglot for subword segmentation and subword weight calculation as the same for the multilingual POS tagging experiment (\secref{sec:pos-tagging}).

We evaluate all the models on multilingual WordSim353 (mWS) and SemLex999 (mSL) from \citet{leviant2015separated}, which is available for English, German, Italian and Russian.
The dataset also contains the relatedness (rel) and similarity (sim) benchmarks derived from mWS. 

We list the results for multilingual word similarity in \tabref{tab:word-sim-multilingual-results}.

\subsection{POS Tagging}
\label{sec:exp-details-pos}
\tabref{tab:pos-settings} and \tabref{tab:sasaki-settings} show the hyperparameter values used in the POS tagging experiment (\secref{sec:pos-tagging}).
For the prediction model, we use the logistic regression classifier from scikit-learn 0.19.1 with the default settings. 

Following the observation in \citet{sasaki2019subword}, we tune the regularization parameter \code{C} for KVQ-FH for all values $a \times 10^b$ where $a=1,\dots,9$ and $b=-1,0,\dots,4$.
We use the POS tagging accuracy for English as criterion, and choose $\text{\code{C}} = 70$.

\tabref{tab:pos-tagging-results-full} lists some statistics of the datasets used in the POS tagging experiment.
\pbos is able to achieve better accuracy over \bos and KVQ-FH in all languages regardless of their morphological type, OOV rate and number of training instances for POS tagging.

\begin{table*}
    \small
    \centering
    \begin{tabular}{l|l|c c c}
      \hline
      \hline
      \multicolumn{2}{c|}{\multirow{2}{*}{Settings}}        & \multicolumn{3}{c}{Model} \\ 
      \multicolumn{2}{c|}{}                                 & \bos & \pbos & \pbosn \\
      \hline
      \multirow{3}{*}{Subwords}       & \code{min\_len}             & 3 & 1 & 1\\ \cline{2-5}
      &                                \code{max\_len}             & 6 & None & None \\ \cline{2-5}
      &                                \code{word\_boundary}       & True & False & False \\ \cline{2-5}
    
      \hline
      \multirow{5}{*}{Training}      & \code{epochs}                & 50 & 50 & 50 \\ \cline{2-5}
      &                               \code{lr}                    & 1.0 & 1.0 & 1.0 \\ \cline{2-5}
      &                                \code{lr\_decay}             & True & True & True \\ \cline{2-5}
      &                                \code{normalize\_semb}        & False & False & True \\ \cline{2-5}
      &                                \code{prob\_eps}             & 0.01 & 0.01 & 0.01 \\ \cline{2-5}

      \hline
      \hline
    \end{tabular}
    \caption{\small
        Training settings used in word similarity experiment for \bos, \pbos, and \pbosn
    }
    \label{tab:ws-settings}
\end{table*}

\begin{table*}
    \small
    \centering
    \begin{tabular}{l|l|c c c}
      \hline
      \hline
      \multicolumn{2}{c|}{\multirow{2}{*}{Settings}}        & \multicolumn{2}{c}{Model} \\ 
      \multicolumn{2}{c|}{}                                 & \bos & \pbos  \\
      \hline
      \multirow{3}{*}{Subwords}       & \code{min\_len}             & 3 & 1 \\ \cline{2-4}
      &                                \code{max\_len}             & 6 & None \\ \cline{2-4}
      &                                \code{word\_boundary}       & True & False  \\ \cline{2-4}
    
      \hline
      \multirow{4}{*}{Training}      &\code{epochs}                & 20 & 20 \\ \cline{2-4}
      &                               \code{lr}                    & 1.0 & 1.0  \\ \cline{2-4}
      &                                \code{lr\_decay}             & True & True  \\ \cline{2-4}
      &                                \code{prob\_eps}             & 0.01 & 0.01  \\ \cline{2-4}

      \hline
      \multirow{1}{*}{Evaluation}    & \code{C}                     & 1 & 1 \\ \cline{2-4}
      
      \hline
      \hline
    \end{tabular}
    \caption{\small
        Training settings used in POS tagging experiment for \bos and \pbos
    }
    \label{tab:pos-settings}
\end{table*}

\begin{table*}
    \small
    \centering
    \begin{tabular}{l|l|c c}
      \hline
      \hline
      \multicolumn{2}{c|}{\multirow{2}{*}{Settings}}        & \multicolumn{2}{c}{Experiment} \\ 
      \multicolumn{2}{c|}{}                                 & Word similarity & POS tagging \\
      
      \hline
      \multirow{3}{*}{Subwords}       & \code{min\_len}             & 3  & 3  \\ \cline{2-4}
      &                                \code{max\_len}             & 30 & 30 \\ \cline{2-4}
      &                                \code{word\_boundary}       & True & True  \\ \cline{2-4}
    
      \hline
      \multirow{3}{*}{Training}      &\code{epochs}                & 300 & 300 \\ \cline{2-4}
      &                                \code{limit\_size}           & 500,000 & 500,000 \\ \cline{2-4}
      &                                \code{bucket\_size}          & 40,000 & 40,000 \\ \cline{2-4}
      
      \hline
      \multirow{1}{*}{Evaluation}    & \code{C}                     & N/A & 70 \\ \cline{2-4}
      
      \hline
      \hline
    \end{tabular}
    \caption{\small
        Training settings used in experiments for KVQ-FH.
    }
    \label{tab:sasaki-settings}
\end{table*}

\begin{table*}
  \small
  \centering
  \begin{tabular}{l | r | r r r | r r | r r}
    \hline\hline
    \multirow{2}{*}{Model}  & \multirow{2}{*}{\# Param}  & \multicolumn{3}{c|}{Dataset} & \multicolumn{2}{c|}{Training Time} & \multicolumn{2}{c}{Prediction Time} \\
    & & WS & RW & Card & Total & Per epoch  & Total & Per word\\
    \hline
    \multicolumn{6}{l}{Target: Polyglot} \\
    \hline
    \bos   & 29.8M  & 34           & \textbf{34} & 6                 & 505s  & 10.1s & 1.9s & 161$\mu$s \\
    KVQ-FH & 7.8M   & 31           & 32          & 12               & 2,669s   & 8.9s  & --     & -- \\
    \pbos  & 37.8M  & \textbf{41}  & 25          & \textbf{15}      & 966s     & 19.3s & 4.2s & 365$\mu$s \\
    \hline
    \multicolumn{6}{l}{Target: Google News} \\
    \hline
    \bos   & 162.7M & 61             & 48          & 11          & 1,110s  & 22.2s & 4.8s & 414$\mu$s\\
    KVQ-FH & 36.2M  & 64             & \textbf{49}          & 21 & 10,638s & 35.5s  & --     & -- \\
    \pbos  & 315.7M & \textbf{68}    & \textbf{49} & \textbf{25}          & 2,065s  & 41.3s & 6.8s & 590$\mu$s\\
    \hline\hline
    
  \end{tabular}
  \caption{\small
    Word similarity performance of subword-level models measured in Spearman's $\rho \times 100$, along with training and prediction time. 
  }
  \label{tab:word-sim-results-full}
\end{table*}

\begin{table*}
    \centering
    \begin{minipage}[b]{0.45\linewidth}
        \small
        \centering
        \begin{tabular}{ c | r r r r }
            \hline\hline
            & mWS & mWS-rel & mWS-sim & mSL \\
            \hline
            \multicolumn{4}{l}{English: 10k tokens $\times$ 300 dim} \\
            \hline
             IV pairs &     65    & 56   &  71   & 26    \\
             All pairs &    29    & 36   &  24   &  7    \\
             OOV &          27\%  & 23\% &  30\% & 36\%  \\
            \hline
            
            \multicolumn{4}{l}{Germen: 10k tokens $\times$ 300 dim} \\
            \hline
             IV pairs &     58   & 50   &  60   & 35    \\
             All pairs &     8   & 14   &   7   &  7    \\
             OOV &          54\% & 52\% &  55\% & 67\%  \\
            \hline
            
            \multicolumn{4}{l}{Italian: 10k tokens $\times$ 300 dim} \\
            \hline
             IV pairs &     52    & 50   &  54   & 24    \\
             All pairs &    11    & 20   &   8   &  2    \\
             OOV &          48\%  & 45\% &  50\% & 54\%  \\
            \hline
            
            \multicolumn{4}{l}{Russian: 10k tokens $\times$ 300 dim} \\
            \hline
             IV pairs &     47    & 32   &  48   & 12    \\
             All pairs &     1    &  4   &   2   &  9    \\
             OOV &          73\%  & 69\% &  75\% & 84\%  \\
            
            \hline\hline
        \end{tabular}
        \caption{\small
            Multilingual target vectors statistics and word similarity performance measured in Spearman's $\rho \times 100$.
        }
        \label{tab:word-sim-multilingual-target-stats}
    \end{minipage}
    \hspace{0.5cm}
    \begin{minipage}[b]{0.45\linewidth}
        \small
        \centering
        \begin{tabular}{l | r | r r r r }
            \hline\hline
            \multirow{2}{*}{Model} & \multirow{2}{*}{\# Param} & \multirow{2}{*}{mWS} & mWS & mWS & \multirow{2}{*}{mSL} \\
              &  & & rel & sim &  \\
            \hline

            \multicolumn{6}{l}{English} \\
            \hline
            \bos   & 20.2M  & 32 & 29 & 34 & 23 \\
            KVQ-FH & 36.0M  & 36 & 41 & 34 & 13 \\
            \pbos  & 30.4M  & 53 & 44 & 61 & 22 \\ 
            \hline

            \multicolumn{6}{l}{Germen} \\
            \hline
            \bos   & 21.3M  & 32 & 24 & 37 & 13 \\
            KVQ-FH & 36.0M  & 18 & 19 & 19 & 14 \\
            \pbos  & 45.8M  & 38 & 30 & 38 & 12 \\ 
            \hline
            
            \multicolumn{6}{l}{Italian} \\
            \hline
            \bos   & 18.8M  & 8 & -2 & 17 & 25\\
            KVQ-FH & 36.0M  & 19 & 22 & 21 & 9 \\
            \pbos  & 35.7M  & 25 & 16 & 27 & 13 \\ 
            \hline
            
            \multicolumn{6}{l}{Russian} \\
            \hline
            \bos   & 20.0M  & 20 & 15 & 21 & 14\\
            KVQ-FH & 36.0M  & 19 & 11 & 24 & 9\\
            \pbos  & 35.6M  & 18 & 12 & 22 & 12 \\ 
            
            \hline\hline
        \end{tabular}
        \caption{\small
            Multilingual word similarity performance of subword-level models measured in Spearman's $\rho \times 100$.
        }
        \label{tab:word-sim-multilingual-results}
    \end{minipage}
\end{table*}

\begin{table*}[!t]
  \small
  \centering
  \begin{tabular}{ l | l | r | r | c c c }
    \hline\hline
  \multirow{2}{*}{Language} & {Morphological} & \multirow{2}{*}{OOV \%} & \multirow{2}{*}{$N_{train}$} & \multicolumn{3}{c}{Model}  \\
            &    Type     &    &     & KVQ-FH & BoS   & \pbos \\
  \hline

Arabic     & Fusional      & 27.1\% & 225,853   & 0.813 & 0.754 & \textbf{0.905} \\
Basque     & Agglutinative & 39.2\% & 72,974    & 0.749 & 0.829 & \textbf{0.866} \\
Bulgarian  & Fusional      & 33.7\% & 50,000    & 0.777 & 0.793 & \textbf{0.929} \\
Chinese    & Isolating     & 70.8\% & 98,608    & 0.633 & 0.330 & \textbf{0.833} \\
Czech      & Fusional      & 58.5\% & 1,173,282 & 0.799 & 0.823 & \textbf{0.917} \\
Danish     & Fusional      & 33.3\% & 88,980    & 0.801 & 0.757 & \textbf{0.904} \\
English    & Analytic      & 26.2\% & 204,587   & 0.770 & 0.770 & \textbf{0.896} \\
Greek      & Fusional      & 18.5\% & 47,449    & 0.866 & 0.888 & \textbf{0.941} \\
Hebrew     & Fusional      & 20.3\% & 135,496   & 0.775 & 0.703 & \textbf{0.915} \\
Hindi      & Fusional      & 27.1\% & 281,057   & 0.811 & 0.800 & \textbf{0.901} \\
Hungarian  & Agglutinative & 29.2\% & 33,017    & 0.777 & 0.794 & \textbf{0.893} \\
Indonesian & Agglutinative & 20.0\% & 97,531    & 0.776 & 0.828 & \textbf{0.899} \\
Italian    & Fusional      & 24.3\% & 289,440   & 0.794 & 0.787 & \textbf{0.930} \\
Kazakh     & Agglutinative & 22.8\% & 4,949     & 0.623 & 0.753 & \textbf{0.815} \\
Latvian    & Fusional      & 23.7\% & 13,781    & 0.722 & 0.756 & \textbf{0.848} \\
Persian    & Agglutinative & 16.9\% & 121,064   & 0.869 & 0.782 & \textbf{0.924} \\
Romanian   & Fusional      & 29.4\% & 163,262   & 0.774 & 0.755 & \textbf{0.898} \\
Russian    & Fusional      & 31.3\% & 79,772    & 0.775 & 0.838 & \textbf{0.911} \\
Spanish    & Fusional      & 29.1\% & 382,436   & 0.818 & 0.769 & \textbf{0.920} \\
Swedish    & Analytic      & 37.4\% & 66,645    & 0.826 & 0.840 & \textbf{0.920} \\
Tamil      & Agglutinative & 28.4\% & 6,329     & 0.702 & \textbf{0.758} & 0.755 \\
Turkish    & Agglutinative & 37.8\% & 41,748    & 0.760 & 0.777 & \textbf{0.837} \\
Vietnamese & Analytic      & 63.8\% & 31,800    & 0.663 & 0.712 & \textbf{0.832} \\
  \hline\hline
  \end{tabular}
  \caption{\small
    Statistics for the languages used in POS tagging experiment.
  }
  $N_{train}$ is the number of training instances for the POS tagging model. 
  OOV \% is the percentage of the words in the POS tagging testing set that is out of the vocabulary of the Polyglot vectors in that language. 
  Experimental results are included for convenience.
  \label{tab:pos-tagging-results-full}
\end{table*}

\end{document}